\documentclass{article}
\usepackage{spconf,amsmath,amsfonts,graphicx,hyperref}
\usepackage{enumitem}
\usepackage[font=small,labelfont=bf]{caption}
\usepackage{subcaption}
\usepackage{booktabs}   
\usepackage{pifont}
\newcommand{\cmark}{\ding{51}}%
\newcommand{\xmark}{\ding{55}}%
\usepackage{amsthm}

\title{Self-Attention Decomposition for Training Free Diffusion Editing}
\name{Tharun Anand\textsuperscript{1}, 
      Mohammad Hassan Vali\textsuperscript{2}, 
      Arno Solin\textsuperscript{2}, 
      Green Rosh\textsuperscript{1}, B H Pawan Prasad\textsuperscript{1}}

\address{\textsuperscript{1} Samsung R\&D Institute India Bangalore \\ 
         \textsuperscript{2} ELLIS Institute Finland \&  Department of Computer Science, Aalto University, Finland}

\begin{document}
\maketitle

\begin{abstract}
Diffusion models achieve remarkable fidelity in image synthesis, yet precise control over their outputs for targeted editing remains challenging. A key step toward controllability is to identify interpretable directions in the model's latent representations that correspond to semantic attributes. Existing approaches for finding interpretable directions typically rely on sampling large sets of images or training auxiliary networks, which limits efficiency. We propose an analytical method that derives semantic editing directions directly from the pretrained parameters of diffusion models, requiring neither additional data nor fine-tuning. Our insight is that self-attention weight matrices encode rich structural information about the data distribution learned during training. By computing the eigenvectors of these weight matrices, we obtain robust and interpretable editing directions. Experiments demonstrate that our method produces high-quality edits across multiple datasets while reducing editing time by 60\% over current benchmarks.
\end{abstract}

\begin{figure*}[htbp]
  \centering
\includegraphics[width=\textwidth,height=0.6\textheight,keepaspectratio]{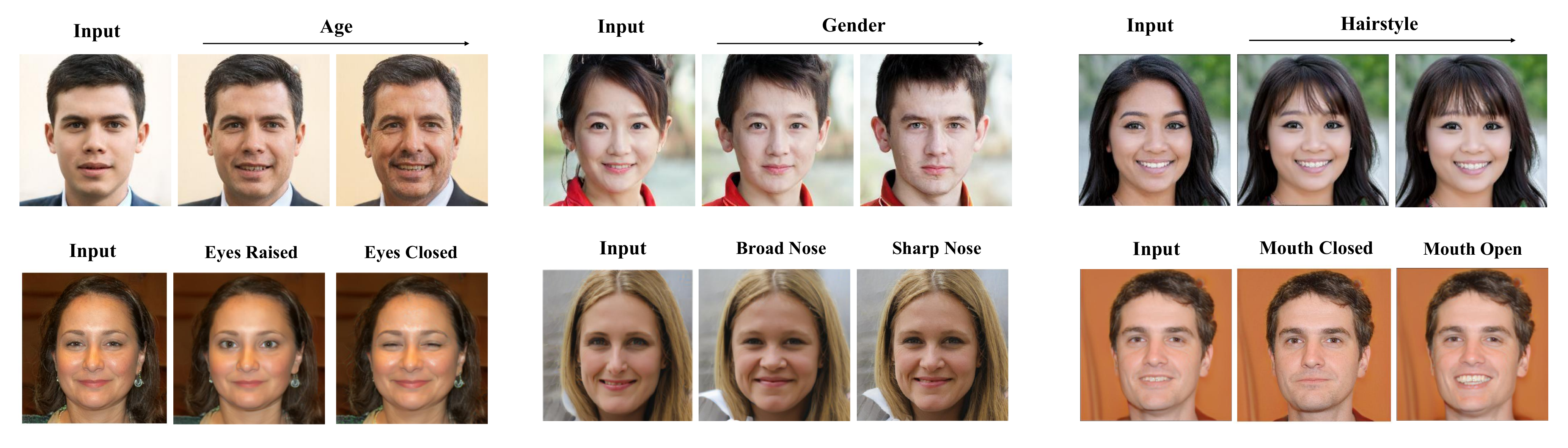}\\[-1em]
  \caption{Illustration of editing capabilities with our method. 
  The first row demonstrates the linear editing property as the perturbation strength \(\alpha\) is varied.  
  The second row highlights the ability of our method to perform precise edits to facial attributes.}
  \label{fig:figure1}
\end{figure*}


\section{Introduction}
\label{sec:intro}
Diffusion models~\cite{ho2020denoising} have emerged as the state-of-the-art for generative modeling, producing diverse, high-fidelity images by iteratively denoising random noise. A crucial application is to control their internal representations to support inference-time tasks such as semantic image editing. However, unlike single-pass generators, the multi-step, iterative nature of the diffusion process results in a unstructured latent space, where semantics are difficult to disentangle. To address this, prior works have explored various techniques for identifying interpretable editing directions. Some methods focus on semantic information within bottleneck features (e.g., the \textit{h-space} framework)~\cite{li2024self}, while others operate directly on the noise latent space~\cite{chen2024exploringlowdimensionalsubspacesdiffusion, zhang2025emergence}. Some others employ techniques like Riemannian geometry~\cite{park2023understanding} and contrastive learning~\cite{dalva2024noiseclr} to find robust editing vectors. However, these approaches face two key limitations. First, they typically rely on sampling a large number of images from the latent space, causing the discovered directions to be inherently tied to the sampled data distribution.
This dependency can introduce a bias and limit the generalization to unseen images. Second, discovering such directions often involves computationally expensive operations~\cite{chen2024exploringlowdimensionalsubspacesdiffusion,park2023understanding} or additional model training~\cite{dalva2024noiseclr,li2024self}.
These limitations substantially increase the editing time.

To address these limitations, we propose a training-free method to discover interpretable semantic directions directly from the parameters of pretrained diffusion models. Motivated by the closed-form factorization technique~\cite{shen2021closed} in GANs that derive semantic directions from linear weight matrices, we extend this paradigm to the more complex architecture of diffusion models, an area that has been largely unexplored. We address this gap by examining the pretrained \textit{Query}, \textit{Key}, and \textit{Value} projection matrices of the self-attention layers. Our core hypothesis is that the self-attention layers of the diffusion models, which are known to encode rich structural and semantic information~\cite{ tumanyan2023plug, cao2023masactrl}, can be utilized to extract interpretable editing directions. Hence, we perform eigen-decomposition on the pretrained query (Q), key (K), and value (V) projection matrices within these layers. The computed eigenvectors serve as robust, sample-independent editing directions that align with prominent data attributes such as age, gender, or smile (for a pretrained face model).



In summary, this paper makes these key contributions:
\begin{enumerate}[itemsep=0pt,parsep=2pt,topsep=2pt,partopsep=0pt,leftmargin=*]
  \item To the best of the author's knowledge, we are the first to derive interpretable editing directions directly from the pretrained weights of diffusion models. We achieve this through an eigen-analysis of the self-attention weights, extracting eigenvectors without any additional training.
  \item We provide a theoretical derivation based on sensitivity of the self-attention operation, demonstrating that principal components of the combined query-key-value weight matrices yield semantically meaningful directions.
  \item We empirically validate our method on facial attribute editing covering gender, age, and expression.  Our results demonstrate that our sample-independent directions generalize effectively across different images, achieving significant reduction in latency.
\end{enumerate}
\section{Proposed Method}
\label{sec:proposed_method}


\noindent\textbf{Problem Formulation:} Diffusion models typically employ a U-Net backbone with self-attention layers to capture long-range dependencies during denoising. At timestep \(t\), let \(Z_t \in \mathbb{R}^{H\times W\times C}\) denote the latent feature map entering a self-attention block. We reshape \(Z_t\) into \(N=H\,W\) tokens of dimension \(d\ = C\), forming \(Z_t \in \mathbb{R}^{N\times d}\). For brevity, we drop the timestep subscript and refer to this matrix simply as \(Z\).

Within a self-attention block, three pretrained projection matrices of $W_Q, \: W_K, \: W_V \in \mathbb{R}^{d\times d}$
map the input representation \(Z\) into queries, keys, and values as
\begin{equation}
Q = Z W_Q, \qquad K = Z W_K, \qquad V = Z W_V \;.
\label{eq:qkv}
\end{equation}
We further decompose the attention computation into intermediate components of
\begin{equation}
L = \frac{Q K^{\!\top}}{\sqrt{d}}, \quad
S = \mathrm{softmax}(L), \quad
\mathrm{Attn}(Z) = S V \;.
\label{eq:attn}
\end{equation}
To uncover meaningful editing directions, we analyze the effect of perturbing the latent features along a candidate direction \(n \in \mathbb{R}^{N\times d}\). Specifically, we introduce a perturbation:
\begin{equation}
 Z' = Z + \alpha\,n, 
 \label{eq:attn}
\end{equation}
Let $\alpha\ll1$ and $n\in\mathbb{R}^{N\times d}$ with $n_i$ its $i$-th row, $\|n_i\|_2=1$.  
The change in the attention output is:
\begin{equation}
\Delta\mathrm{Attn} = \mathrm{Attn}(Z') - \mathrm{Attn}(Z).
\label{eq:z}
\end{equation}
Our objective is to derive directions $n^*$ that maximize the impact on the attention output, thus directly correspond to axes of semantic variation \cite{tzelepis2023sefa}:
\vspace{2 mm}
\begin{equation}
n^* = \arg \max_{n \in \mathbb{R}^{N \times d}, \|n_i\|_2= 1} \|\Delta\mathrm{Attn}\|_2^2 .
\label{eq:n*}
\end{equation}

\noindent\textbf{Latent Whitening: }
Our derivation is based on the observation that U-Net intermediate representations during early diffusion denoising steps are approximately whitened, i.e., their feature covariances are close to identity. This extends prior empirical observations of \emph{latent whitening}, where diffusion latents exhibit near-isotropic statistics due to injected Gaussian noise, particularly at high-noise timesteps~\cite{everaert2024covariancemismatch,biroli2024dynamical}. In this regime, which we empirically target ($0.8T \geq t \geq 0.5T$), the covariance structure of intermediate features becomes approximately diagonal, enabling the analytical simplification
\begin{equation}
\mathbb{E}\!\left[Z^{\top} Z\right] \approx I.
\end{equation}
This isolates self-attention effects from the data distribution, enabling a closed-form eigen decomposition with semantically coherent directions across datasets and architectures.

\label{whitening_approximation}

\subsection{Weight Space Eigen Decomposition}
\label{sec:eigenanalysis}
Given a small perturbation to the input latents:
\begin{equation}
Z' = Z + \alpha n, \qquad \|n_i\|_2 = 1, \qquad \alpha \text{ is small},
\label{eq:z}
\end{equation}
%
This induces perturbations in attention projections:
\begin{equation}
\Delta Q = \alpha nW_Q, \quad
\Delta K = \alpha nW_K, \quad
\Delta V = \alpha nW_V
\label{eq:attn}
\end{equation}
To calculate the change in score matrix $\Delta L$ induced by the perturbation, while neglecting the \(\mathcal{O}(\alpha^2)\) term:
\begin{align}
\Delta L
   &= \frac{1}{\sqrt{d}}
      \bigl(
        Q\,\Delta K^{\!\top}
        + \Delta Q\,K^{\!\top}
      \bigr) .
\label{eq:deltaL}
\end{align}
Applying the first-order Taylor expansion of the softmax function S to obtain $\Delta S$:
\begin{align}
\Delta S
   &= J_{\mathrm{softmax}}(L)\,\Delta L ,
\label{eq:deltaS}
\end{align}
where \(J_{\mathrm{softmax}}(L)\) is the Jacobian of softmax at \(L\). 
Now, since \(\mathrm{Attn}(Z)=S V\),  $\Delta \mathrm{Attn}$ decomposes as:
\begin{align}
\Delta \mathrm{Attn}
   &= \Delta S\,V + S\,\Delta V
\label{eq:deltaAttn}
\end{align}

\begin{figure*}[t!]
  \centering
  \includegraphics[
    width=1.0\textwidth,      
    height=0.6\textheight,    
    keepaspectratio
  ]{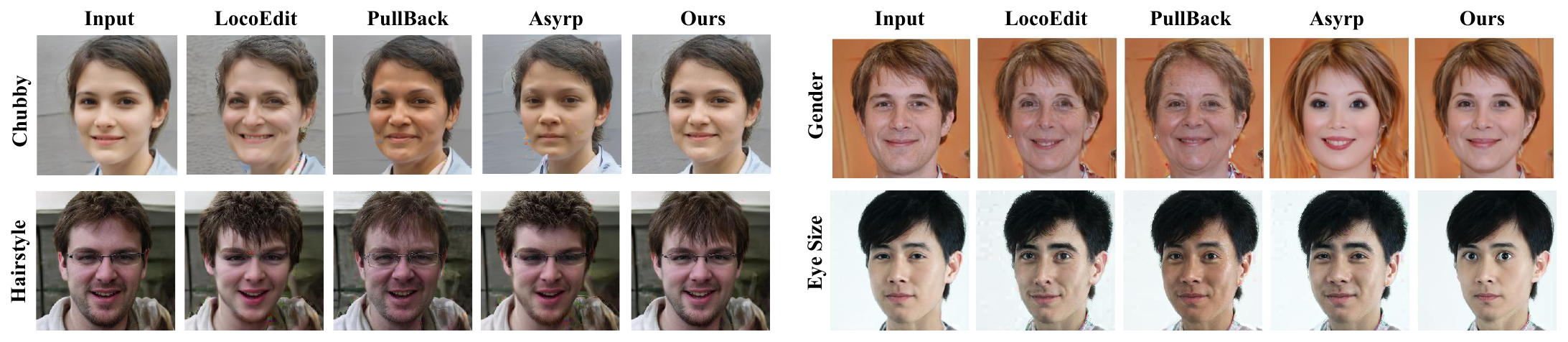}\\[-4pt]
  \caption{Our method supports a diverse range of edits with improved disentanglement, enabling precise single-step manipulation of the target attribute beyond existing benchmarks.}
  \label{fig:sota}
\end{figure*}

\noindent\textbf{Objective Function:} 
Maximize $\Delta \mathrm{Attn}$ as defined in Eq.~\eqref{eq:n*}.
\begin{equation}
\|\Delta \mathrm{Attn}\|_F^2 = \|\Delta S\,V + S\,\Delta V\|_F^2.
\label{eq:norm}
\end{equation}
Upon simplification:
\begin{align}
\bigl\|\Delta S\,V + S\,\Delta V\bigr\|_F^2
   &= \bigl\|\Delta S\,V\bigr\|_F^2
      + \bigl\|S\,\Delta V\bigr\|_F^2 \nonumber \\
   &\quad
      + 2\,\mathrm{tr}\!\bigl((\Delta S\,V)^{\top}(S\,\Delta V)\bigr).
\end{align}

\noindent Under the whitening approximation(Sec.~\ref{whitening_approximation}), perturbations to S and V become weakly correlated, allowing us to neglect the cross term in expectation.\sloppy
\begin{align}
\mathbb{E}\|\Delta \mathrm{Attn}\|_F^2
\approx \mathbb{E}\|\Delta S\,V\|_F^2 + \mathbb{E}\|S\,\Delta V\|_F^2.
\label{eq:expectation}
\end{align}

\noindent\textbf{Term 1:} For the first term, we utilize the cyclic property of the trace of a matrix:
\begin{equation}
\|\Delta S\, V\|_F^2 = \text{tr}(V^\top (\Delta S)^\top \Delta S\, V) = \text{tr}( (\Delta S)^\top \Delta S\, V.V^\top) .
\end{equation}
From Eqs.~\eqref{eq:deltaL} and~\eqref{eq:deltaS}, we have $\Delta S$:
\begin{align}
\Delta S = \frac{\alpha}{\sqrt{\mathrm{d}}}\,
\bigl[ Q W_K^\top n + K W_Q^\top n \bigr].
\label{eq:expectation}
\end{align}
%
%
With approximation (\ref{whitening_approximation}), the latent-dependent covariance reduces to identity ($\mathbb{E}[V V^{\top}] \approx I$), isolating weight matrices:
\begin{align}
\mathbb{E}[\|\Delta S\, V\|_F^2] \approx \alpha^2 n^{\top} (W_Q^{\top} W_Q + W_K^{\top} W_K) n .
\end{align}

\noindent\textbf{Term 2:}
\vspace{-1 mm}
\begin{align}
\|S\, \Delta V\|_F^2 = \text{tr}((\Delta V)^\top S^\top S\, \Delta V) .
\label{eq:Term2}
\end{align}
Since $\mathbb{E}[S^{\top} S] \approx  I $, (Sec.~\ref{whitening_approximation}). The term then reduces to 
\begin{align}
\mathbb{E}[\|S\, \Delta V\|_F^2] \approx \alpha^2 n^{\top} W_V W_V^{\top} n.
\label{eq:E[term2]}
\end{align}

\noindent\textbf{Final Expression:} 
Combining terms 1 \& 2 yields,
\begin{equation}
\mathbb{E}\bigl\|\Delta \mathrm{Attn}\bigr\|_F^2
   \approx
   \alpha^2 n^{\top}
   \Bigl(W_Q^{\top}W_Q + W_K^{\top}W_K + W_V W_V^{\top}\Bigr) n .
\end{equation}
Define the combined matrix:
\begin{align}
C = W_{Q}^{\top}W_{Q} + W_{K}^{\top}W_{K} + W_{V}^{\top}W_{V}.
\label{eq:C}
\end{align}
Now maximizing $n^{\top}C\,n$ subject to $\|n\|_{2} = 1$ is a standard Rayleigh quotient problem. The optimal direction $n^{*}$ is the principal eigenvector of $C$. Subsequent eigenvectors provide orthogonal editing directions. As shown in prior work \cite{kwon2023diffusion,chen2024exploringlowdimensionalsubspacesdiffusion}, editing vectors are most effective when applied at early 
denoising timesteps, before image content forms, where the whitening assumption 
$\mathbb{E}[ZZ^\top] \approx I$ holds. Following a similar strategy to prior works, we empirically choose   
the range $0.5T < t < 0.8T$.
\begin{equation}
z' =
\begin{cases}
z + \alpha n^*,
 & \text{if } 0.5T < t < 0.8T, \\
z, & \text{otherwise}.
\end{cases}
\end{equation}

\vspace{-0.5 mm}
\section{Experiments and Results}
\noindent We evaluate the effectiveness of our proposed method using pretrained DDPM~\cite{ho2020denoising} models on CelebA-HQ(Faces)~\cite{liu2015faceattributes,karras2018progressive}, LSUN Cats, Cars and Rooms~\cite{yu2015lsun}, and LSUN Cars, ensuring coverage across multiple datasets. We evaluate our method against four existing benchmarks: Asyrp~\cite{kwon2023diffusion}, Locoedit~\cite{chen2024exploringlowdimensionalsubspacesdiffusion}, Pullback~\cite{park2023understanding}, and NoiseCLR~\cite{dalva2024noiseclr}.
\vspace{2 mm}

\noindent\textbf{Implementation Details:} All experiments use unconditional, pretrained diffusion model with \(T = 1000\) total denoising timesteps. Unconditional diffusion models employ U-Nets with a self-attention layer in both the encoding and decoding blocks. In our work, we utilize the self-attention layer from the encoding block~\cite{lin2024ctrlx,si2024freeu} to apply our editing. All experiments are executed on a single NVIDIA V100 GPU (32~GB).


\vspace{-2 mm}

\subsection{Qualitative Assessment}
\noindent We highlight key qualitative properties of our weight-space eigenanalysis in Fig.~\ref{fig:figure1}:

\smallskip

\noindent \textbf{Linearity:} We exploit locally linear subspaces in the attention latents, so that linear shifts in latents produce proportionally scaled semantic changes, unlike prior methods that rely on nonlinear subspaces for editing \cite{park2023understanding,kwon2023diffusion} (Fig.~\ref{fig:figure1}, row 1).

\smallskip

\begin{table*}[!t]
  \centering
  \footnotesize
  \renewcommand{\arraystretch}{1.2}
  \setlength{\tabcolsep}{6pt}
  \caption{Comparison of performance metrics across editing methods.}
  \label{tab:qualitative_properties}
  \begin{tabular}{lccccc}
    \toprule
    \textbf{Method Name} & \textbf{Editing Time} & \textbf{Sample Independent Directions} & \textbf{Localized Edit?} & \textbf{Supervision Required?} & \textbf{Images for Learning} \\
    \midrule
    Pullback~\cite{park2023understanding}          & 80~s   & \xmark & \xmark & \cmark & 20   \\
    NoiseCLR~\cite{dalva2024noiseclr}          & 1~day & \xmark & \xmark & \cmark & 100 \\
    Asyrp~\cite{kwon2023diffusion}             & 475~s & \xmark & \xmark & \cmark & 100 \\
    LOCOEdit~\cite{chen2024exploringlowdimensionalsubspacesdiffusion}          & 79~s  & \xmark & \cmark & \xmark &   1\\
    \midrule
    \textbf{Ours} & 3~s & \cmark & \cmark & \xmark & 0   \\
    \bottomrule
  \end{tabular}
\end{table*}

\noindent \textbf{Diverse Editing Directions:} Our analysis shows that
the discovered eigenvectors capture a spectrum of editing
directions in the latent space. The leading eigenvectors
correspond to global semantic attributes like age
and gender, while subsequent directions progressively
focus on finer, localized regions of the face.(Fig.~\ref{fig:figure1}, row.~2).

\vspace{1 mm}
 \begin{figure}[t!]
  \raggedright
\includegraphics[width=\columnwidth,trim=0 0 20 0,clip]{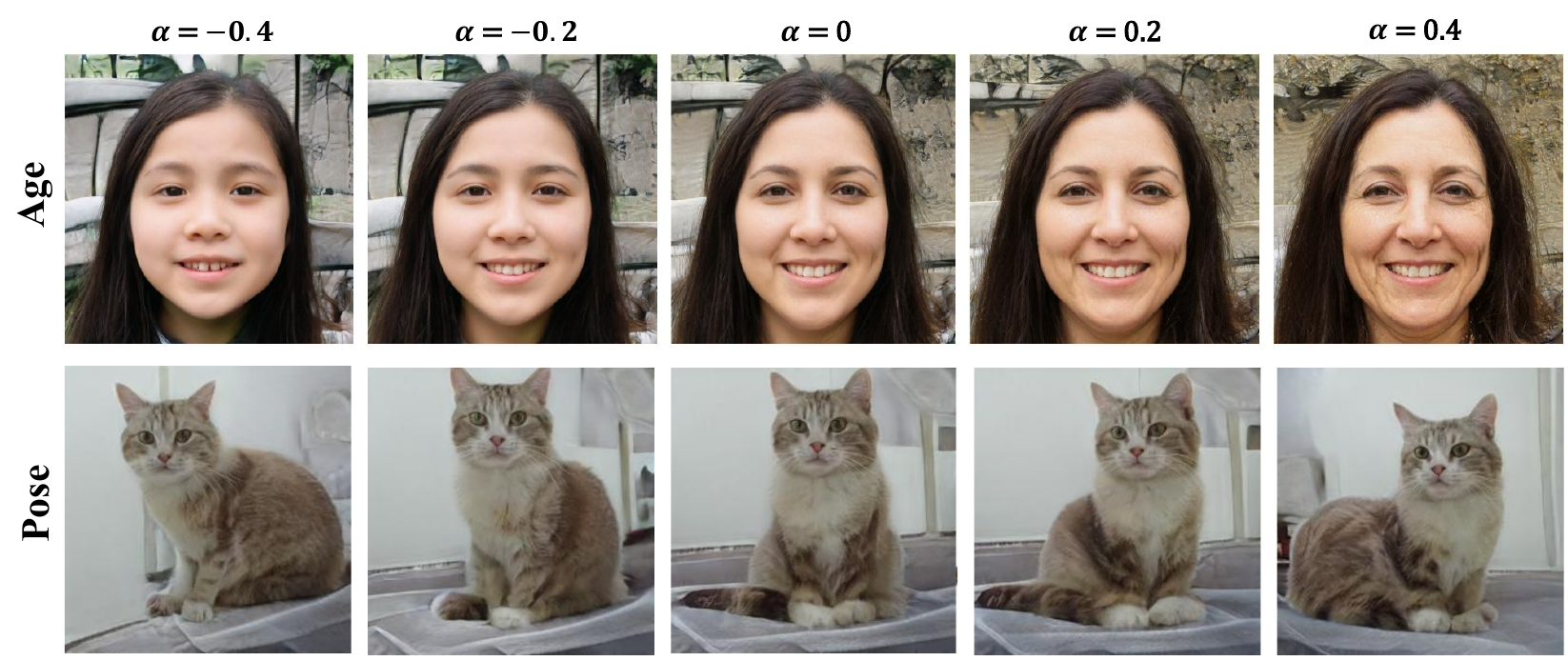}
  \caption{Illustration of how the edit strength parameter $\alpha$ controls the linear change in attribute intensity.}
  \label{fig:ablation}
\end{figure}

\noindent Figure~\ref{fig:sota} shows qualitative comparisons with benchmarks, demonstrating that our method edits diverse facial attributes while better preserving unedited regions than the baselines.


\vspace{-2 mm}
\subsection{Quantitative Analysis}
In our quantitative analysis, we measure both image‐level fidelity and semantic alignment to the intended edits. We measure pixel- and structure-level consistency using SSIM and PSNR, identity preservation via FaceNet~\cite{7298682} cosine similarity, and semantic alignment with the Directional CLIP score~\cite{Kim_2022_CVPR}. For evaluation, we test 15 semantic editing directions on 100 image samples, resulting in 1500 test cases. Table~\ref{tab:quantitative_metrics} compares our method and four baselines methods on the CelebA-HQ test set. Comparative results against existing benchmarks for other key evaluation metrics are reported in Table~\ref{tab:qualitative_properties}. Experiments show that our method reduces latency by 60\% while consistently demonstrating superior results across baselines in both semantic alignment and image-level metrics.

\begin{table}[h]
  \centering
  \footnotesize
  \renewcommand{\arraystretch}{1.1}
  \caption[Quantitative analysis on CelebA-HQ]{Quantitative comparison of editing methods on CelebA-HQ. Higher is better for all metrics.}
  \label{tab:quantitative_metrics}
  \begin{tabular}{lcccc}
    \toprule
    \textbf{Method}      & \textbf{SSIM $\uparrow$} & \textbf{PSNR $\uparrow$} & \textbf{ID-Sim $\uparrow$} & \textbf{Dir-CLIP $\uparrow$} \\
    \midrule
    Asyrp~\cite{kwon2023diffusion}              & 0.88            & 25.2 dB         & 0.75              & 0.62                \\
    Pullback~\cite{park2023understanding}           & 0.90            & 26.8 dB         & 0.78              & 0.65                \\
    LocoEdit~\cite{chen2024exploringlowdimensionalsubspacesdiffusion}           & 0.87            & 24.5 dB         & 0.70              & 0.60                \\
    NoiseCLR~\cite{dalva2024noiseclr}          & 0.85            & 23.9 dB         & 0.68              & 0.58                \\
    \midrule
    \textbf{Ours}      & \textbf{0.92}   & \textbf{28.5 dB}& \textbf{0.82}     & \textbf{0.75}       \\
    \bottomrule
  \end{tabular}
\end{table}


\begin{figure}[htbp]
  \centering
  \includegraphics[width=\columnwidth,trim=20 0 20 0,clip]{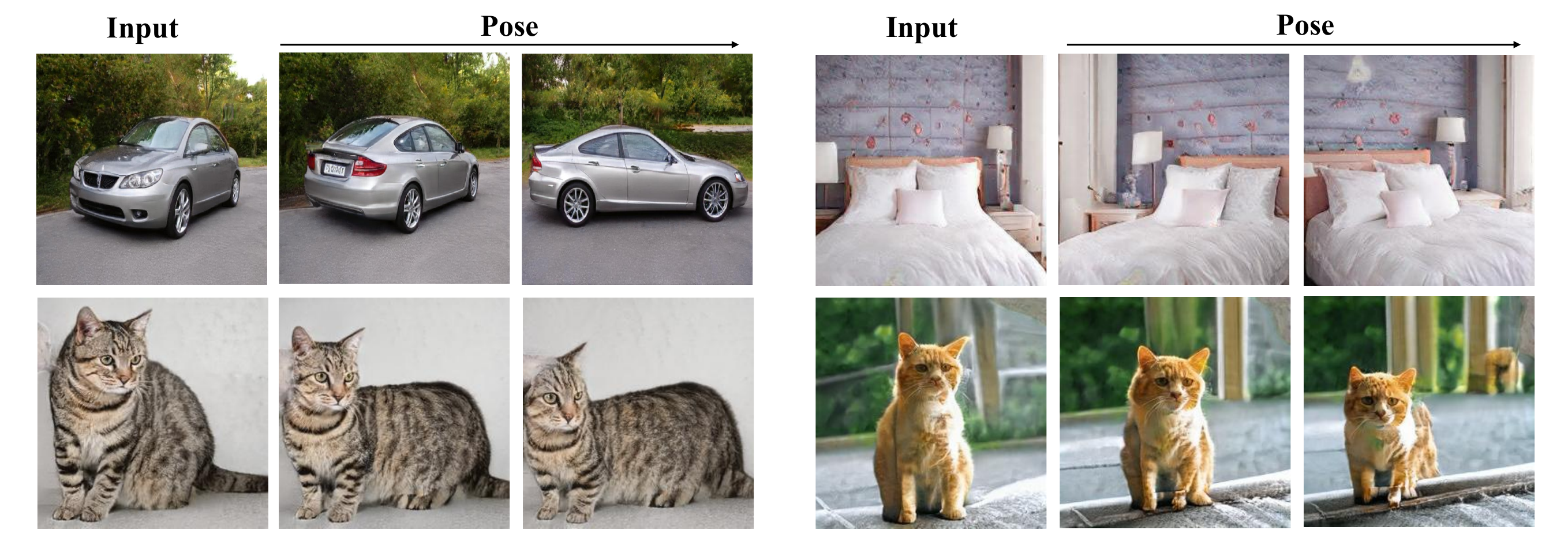}
  \caption{Effect of our proposed editing method on pretrained models across multiple datasets.}
  \label{fig:ablation1}
\end{figure}

 \subsection{Ablation Studies}
 We conduct the following ablations on the following key components in our editing method:

\smallskip


\noindent \textbf{Perturbation Strength:} We sweep $\alpha \in [-0.4, 0.4]$ to visually show the effect on the linear change in the attribute strength of the edit. Fig.~\ref{fig:ablation} shows that the strength of the edit can be directly controlled using $\alpha$.

\smallskip

\noindent \textbf{Editing Across Diverse Datasets:} Our method is universally applicable across models pretrained on various datasets, as demonstrated in Figure~\ref{fig:ablation1}, showcasing its effect on models pretrained on the LSUN datasets~\cite{yu2015lsun} (Cars, Cats, and Rooms). The principal vectors in these models represent variations in body pose and shape, as shown in Figure~\ref{fig:ablation1}.



\section{Conclusion}
In this paper, we introduce a training-free method for identifying editing directions in diffusion models via eigenanalysis of the self-attention weights in pretrained diffusion models. Our theoretical derivation in Sec.~\ref{sec:proposed_method} demonstrates that the principal eigenvectors of the combined query-key-value weight matrices correspond to semantically meaningful editing directions, providing a closed-form solution to the image editing problem in diffusions. Future work includes extending this weight-space analysis to multi-modal diffusion models, where cross-attention layers may reveal joint editing directions. Extending our eigen-analysis to the cross-attention layers of recent DiTs for joint text–visual control is a direction for future work, particularly to study robustness in higher-dimensional latent spaces.

\bibliographystyle{IEEEbib}
\bibliography{strings,refs}

\end{document}